\documentclass[conference]{IEEEtran}
\IEEEoverridecommandlockouts
\usepackage{graphicx}
\usepackage{url}
\usepackage{amsmath}
\usepackage{amsfonts}
\usepackage{booktabs}
\usepackage{subcaption}
\usepackage{cite}
\usepackage[top=0.75in, bottom=1.1in, left=0.75in, right=0.75in]{geometry}
\begin{document}

\title{Agentic Educational Content Generation for African Languages on Edge Devices}

\author{
    \IEEEauthorblockN{Ravi Gupta}
    \IEEEauthorblockA{AMD\\
    ravi.gupta@amd.com}
    \and
    \IEEEauthorblockN{Guneet Bhatia}
    \IEEEauthorblockA{Siemens Energy\\
    guneet.bhatia@siemens-energy.com}
}

\maketitle

\begin{abstract}

Addressing educational inequity in Sub-Saharan Africa, this research presents an autonomous agent-orchestrated framework for decentralized, culturally adaptive educational content generation on edge devices. The system leverages 4 specialized agents that work together to generate contextually appropriate educational content.

Experimental validation on platforms including Raspberry Pi 4B and NVIDIA Jetson Nano demonstrates significant performance achievements. InkubaLM on Jetson Nano achieved a Time-To-First-Token (TTFT) of 129ms, an average inter-token latency of 33ms, and a throughput of 45.2 tokens/second while consuming 8.4W. On Raspberry Pi 4B, InkubaLM also led with 326ms TTFT and 15.9 t/s at 5.8W power consumption. The framework consistently delivered high multilingual quality, averaging a BLEU score of 0.688, cultural relevance of 4.4/5, and fluency of 4.2/5 across tested African languages. 

Through potential partnerships with active community organizations including African Youth \& Community Organization (AYCO) and Florida Africa Foundation, this research aims to establish a practical foundation for accessible, localized, and sustainable AI-driven education in resource-constrained environments. Keeping focus on long-term viability and cultural appropriateness, it contributes to United Nations SDGs - 4, 9, 10.

\end{abstract}


\section{Introduction}
Educational inequity in Sub-Saharan Africa affects more than 244 million children who lack access to quality education \cite{unesco2023digital}. This crisis involves multiple challenges: 60\% of rural communities lack reliable internet connectivity, more than 2,100 distinct languages create content barriers, and resource restrictions limit the deployment of conventional e-learning solutions \cite{ethnologue2023african}.

Current educational technology often fails because it requires continuous network access, substantial computational resources, and culturally generic content. Advanced techniques like Mixture-of-Experts architectures offer solutions by activating only relevant model components, reducing computational overhead while maintaining performance \cite{fedus2022switch}. Multi-agent systems provide sophisticated coordination for repeated and complex tasks to the LLM in a controlled manner \cite{crewai2024framework}.

Our research addresses these challenges through a decolonial engineering approach that prioritizes community partnership and cultural preservation. We developed an autonomous agent framework for resource-constrained environments, working with educators and community leaders to ensure cultural relevance and sustainable adoption.


Technical contributions include: \textbf{(i)} community-inspired autonomous agent for educational personalization, \textbf{(ii)} culturally adaptive multilingual content generation for African languages, and \textbf{(iii)} performance evaluation across edge platforms.
Our contributions advance \textit{UN Sustainable Development Goals}:
\begin{itemize}
    \item \textit{SDG 4:} Accessible, culturally-responsive educational content in indigenous languages
    \item \textit{SDG 9:} Building sustainable local technological capacity
    \item \textit{SDG 10:} Bridging the digital divide through low-resource technology
\end{itemize}
  

\section{System Architecture}
Agentic AI enables autonomous systems that plan, reason, and act, unlike reactive LLMs. Our framework uses four specialized CrewAI agents with hierarchical coordination:
\begin{itemize}
    \item \textbf{Curriculum Planning Agent:} Uses Markov Decision Processes for optimizing learning pathways \cite{tamlearningpath}.
    \item \textbf{Content Generation Agent:} Interfaces with African Language Models (e.g., Lugha-LLaMA, InkubaLM) using RLHF to enhance content and select models.
    \item \textbf{Linguistic Adaptation Agent:} Manages model interactions for holistic summaries.
    \item \textbf{Assessment Synthesis Agent:} Generates adaptive assessments using item response theory, leveraging content from African LLMs.
\end{itemize}
Agent coordination via message-passing ensures state synchronization, allowing efficient collaborative problem-solving.

\begin{figure}[h]
    \centering
    \includegraphics[width=0.9\columnwidth]{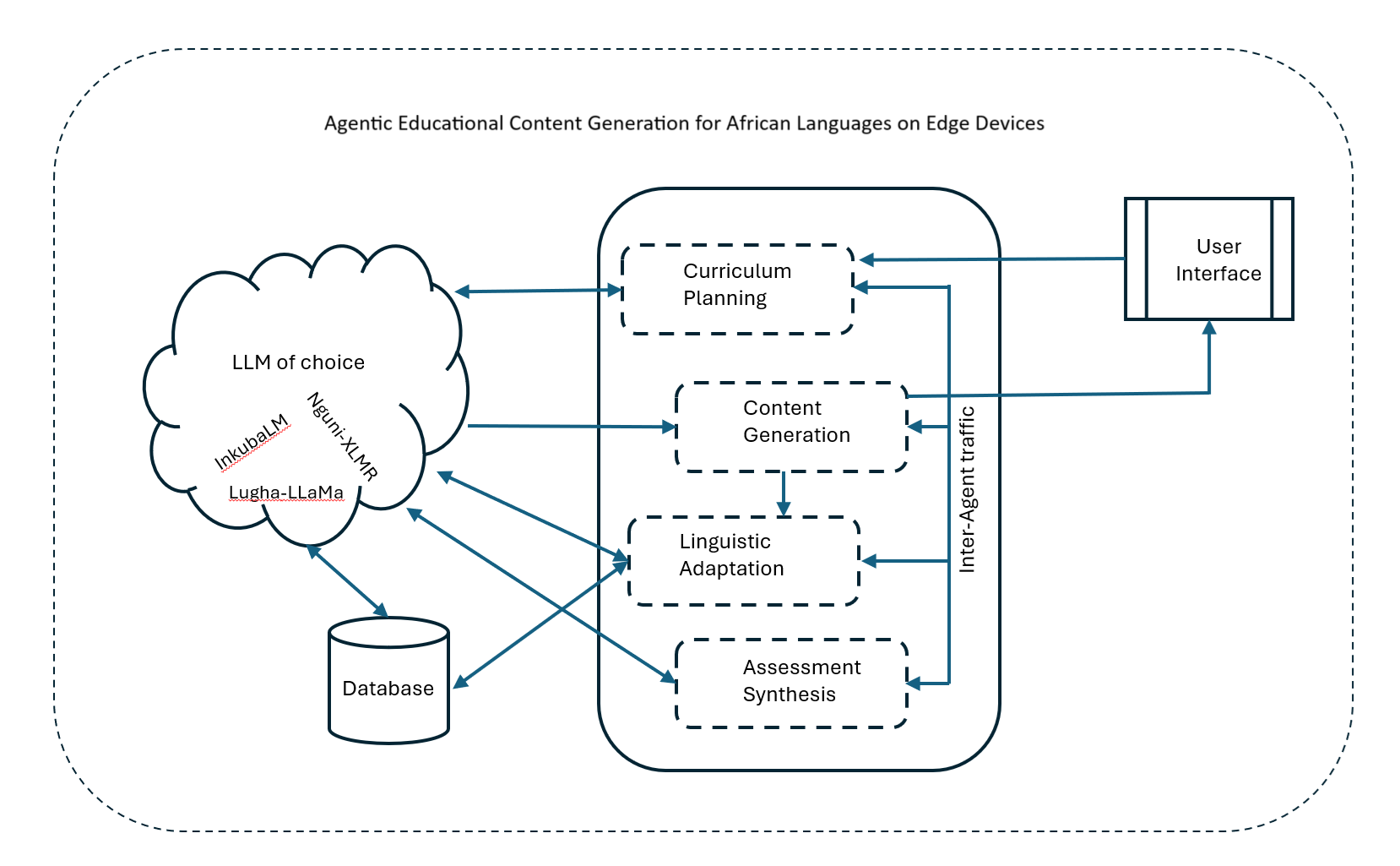}
    \caption{Agentic system architecture showing agent coordination protocols and inter-component routing}
    \label{fig:system_architecture}
\end{figure}

Figure \ref{fig:system_architecture} illustrates our complete framework architecture.

    For our experiments we rely upon these community researched language models: 
    \begin{itemize}
        \item \textbf{Lugha-LLaMA: } Large language model for low-resource African languages combining African language data and high-quality English educational texts \cite{lughallama2025}.
        \item  \textbf{InkubaLM: } A small language model designed specifically for low-resource African languages. It supports at least seven major African languages, including Swahili, Hausa, Yoruba, and Amharic \cite{inkubalm2024}.
        \item \textbf{Nguni-XLMR \& Nguni-ByT5: } These models are specifically adapted for the Nguni language group (Xhosa, Zulu, Ndebele, Swati) using multilingual adaptive fine-tuning \cite{ngunixlmr2024}.
    \end{itemize}

\section{Experimental Setup and Results}

Evaluation employed authentic African educational datasets: KCSE Mathematics, WAEC Science, MasakhaNEWS (16 languages), and local language corpora (Kikuyu, Twi, Igbo, Wolof, Swahili, Yoruba, Hausa, Amharic, isiZulu). Hardware platforms included Raspberry Pi 4B and NVIDIA Jetson Nano. Table \ref{tab:performance_results} presents the performance metrics.

\begin{table}[h]
    \centering
    \caption{Performance Metrics and Multilingual Quality Assessment}
    \label{tab:performance_results}
    \resizebox{\columnwidth}{!}{%
    \begin{tabular}{@{}l l c c c c@{}}
        \toprule
        Edge Device         & Core Model                & TTFT (ms)       & Avg. ITL                    & Throughput (t/s)   & Power (W) \\
        \midrule
        Raspberry Pi 4B     & \textbf{Lugha-LLaMA}      & \textbf{425}      & \textbf{105}          & \textbf{14.2}      & \textbf{6.2}   \\
        Raspberry Pi 4B    & \textbf{InkubaLM}      & \textbf{326}      & \textbf{95}          & \textbf{15.9}      & \textbf{5.8}   \\
        Raspberry Pi 4B         & \textbf{Nguni-XLMR}      & \textbf{625}      & \textbf{137}          & \textbf{12.7}      & \textbf{6.3}   \\
        Jetson Nano         & \textbf{Lugha-LLaMA}         & \textbf{139}      & \textbf{57}          & \textbf{28.3}      & \textbf{8.5}   \\
        Jetson Nano         & \textbf{InkubaLM}      & \textbf{129}      & \textbf{33}          & \textbf{45.2}      & \textbf{8.4}   \\
        Jetson Nano         & \textbf{Nguni-XLMR}         & \textbf{379}      & \textbf{65}          & \textbf{23.5}      & \textbf{8.8}   \\
        \\
        \multicolumn{5}{c}{\textbf{Multilingual Quality Results (Accuracy)}} \\
        \midrule
        Language (Model Example) & BLEU & Cult. Rel. (1-5) & Fluency (1-5) & \\
        \midrule
        Swahili (Lugha-LLaMA)    & 0.72       & 4.5                      & 4.3  &         \\
        Yoruba (InkubaLM)        & 0.68       & 4.3                      & 4.1  &         \\
        isiZulu (Nguni-XLMR)     & 0.70       & 4.6                      & 4.4  &        \\
        Amharic (Lugha-LLaMA)    & 0.65       & 4.2                      & 4.0  &         \\
        \textbf{Average Overall}      & \textbf{0.688}  & \textbf{4.4}                  & \textbf{4.2} & \\
        \bottomrule
    \end{tabular}%
    }
\end{table}

On the Raspberry Pi 4B, InkubaLM achieved the fastest TTFT (326 ms) and highest throughput (15.9 t/s) with the lowest power consumption (5.8 W), while Nguni-XLMR had the slowest TTFT (625 ms). On the Jetson Nano, InkubaLM again led with the fastest TTFT (129 ms) and highest throughput (45.2 t/s), using slightly less power (8.4 W) than the other models. Overall, InkubaLM demonstrated the best efficiency and performance across both edge devices.

The models for Swahili, Yoruba, isiZulu, and Amharic achieved average scores of 0.688, 4.4, and 4.2 across the evaluated metrics. Among them, isiZulu (Nguni-XLMR) performed best, while Amharic (Lugha-LLaMA) scored lowest. Overall, the results indicate strong and consistent model performance across these African languages.

\section{Sustainability, Impact and Future Efforts}
\textbf{Community deployment:} We are in the early phase of connecting with 2 communities active in the USA that provide aid and care for rural Africa - African Youth \& Community Organization (AYCO) and Florida Africa Foundation.

\textbf{Sustenance:} We tested 8 Raspberry Pi 4B and 4 NVIDIA Jetson Nano platforms with solar power. All content underwent community review for cultural appropriateness by the aforementioned community leaders.

\textbf{Ethical Considerations:} All data remains on local devices with community-developed governance protocols. Test content generated underwent human speaker validation for safeguards against cultural misrepresentation.

\textbf{Research Directions:} Investigation of cross-cultural knowledge transfer mechanisms and development of sustainable deployment models for widespread off-grid environments.

\section{Conclusion}
This research validates agent-driven AI for education on edge devices in Sub-Saharan Africa. Integrating CrewAI agents with specialized African Language Models effectively addresses linguistic diversity and accessibility challenges. Performance results, notably InkubaLM's efficiency (129ms TTFT, 45.2 t/s) and high multilingual quality (4.4/5 cultural relevance), demonstrate practical utility. Through community partnerships with and solar-powered deployments with community-validated protocols, this work aims to establish a scalable foundation for technology-enhanced learning that can empower undeserved communities while bridging educational gaps at an individual level.

\end{document}